\documentclass[nohyperref]{article}

\usepackage{microtype}
\usepackage{graphicx}
\usepackage{booktabs} 

\usepackage[hypertexnames=false]{hyperref}

\usepackage{amsmath}
\usepackage{amssymb}
\usepackage{mathtools}
\usepackage{amsthm}

\usepackage{amsmath,amsfonts,bm}

\def\eqref#1{equation~\ref{#1}}

\def\1{\bm{1}}

\DeclareMathAlphabet{\mathsfit}{\encodingdefault}{\sfdefault}{m}{sl}
\SetMathAlphabet{\mathsfit}{bold}{\encodingdefault}{\sfdefault}{bx}{n}

\usepackage{subcaption}
\usepackage{caption}

\usepackage[accepted]{icml2022}

\usepackage[capitalize,noabbrev]{cleveref}

\theoremstyle{plain}

\theoremstyle{definition}

\theoremstyle{remark}

\usepackage[textsize=tiny]{todonotes}

\icmltitlerunning{Towards Galaxy Foundation Models with Hybrid Contrastive Learning}

\begin{document}

\twocolumn[
\icmltitle{Towards Galaxy Foundation Models with Hybrid Contrastive Learning}



\icmlsetsymbol{equal}{*}

\begin{icmlauthorlist}
\icmlauthor{Mike Walmsley}{uom}
\icmlauthor{Inigo Val Slijepcevic}{uom}
\icmlauthor{Micah Bowles}{uom}
\icmlauthor{Anna M. M. Scaife}{uom}
\end{icmlauthorlist}

\icmlaffiliation{uom}{Department of Physics and Astronomy, University of Manchester, Manchester, UK}

\icmlcorrespondingauthor{Mike Walmsley}{michael.walmsley@manchester.ac.uk}

\icmlkeywords{Machine Learning, ICML}

\vskip 0.3in
]



\printAffiliationsAndNotice{\icmlEqualContribution} 

\begin{abstract}
New astronomical tasks are often related to earlier tasks for which labels have already been collected.
We adapt the contrastive framework BYOL to leverage those labels as a pretraining task while also enforcing augmentation invariance.
For large-scale pretraining, we introduce GZ-Evo v0.1, a set of 96.5M volunteer responses for 552k galaxy images plus a further 1.34M comparable unlabelled galaxies.
Most of the 206 GZ-Evo answers are unknown for any given galaxy, and so our pretraining task uses a Dirichlet loss that naturally handles unknown answers. GZ-Evo pretraining, with or without hybrid learning, improves on direct training even with plentiful downstream labels (+4\% accuracy with 44k labels). Our hybrid pretraining/contrastive method further improves downstream accuracy vs. pretraining or contrastive learning, especially in the low-label transfer regime (+6\% accuracy with 750 labels).

\end{abstract}

\section{Introduction}
\label{sec:intro}

Contrastive learning frameworks are typically evaluated by their ability to learn representations useful for downstream tasks given minimal labelled data \citep{Le-Khac2020}. All existing applications of contrastive learning to astronomical images follow this pattern \citep{Hayat2021self, Sarmiento2021, Stein2022}. However, in practice, astronomers need not start from a blank slate.

New astronomical tasks are often related to earlier tasks for which labels have already been collected.
In the context of galaxy morphology (i.e. visual appearance), new versions of the Galaxy Zoo citizen science project \citep{Masters2019} gradually shift the questions asked of volunteers (`does this galaxy have a bar' to `how large is the bar', `is the galaxy merging' to `what kind of merger', etc.). The responses to previous questions are therefore likely to be useful in predicting the answers to new questions.

We propose using the responses to all questions in all previous Galaxy Zoo versions to create generalizable representations useful for answering new questions, in a hybrid approach that jointly combines both contrastive learning and supervised pretraining. Specifically, in this work:

\begin{enumerate}
    \item We assemble the largest and broadest galaxy morphology dataset to date, with 552k labelled and 1.34m unlabelled galaxies from five telescopes and four Galaxy Zoo campaigns
    \item We adapt the contrastive framework BYOL \citep{Grill2020} to additionally calculate a supervised regression loss that forces the representation to solve 206 galaxy morphology tasks while remaining invariant to contrastive augmentations.
\end{enumerate}

We are motivated by CS literature emphasizing the practical value of broad (typically self-supervised) pretraining at scale \cite{Hendrycks2019a, Bommasani2021, Goyal2021, Azizi2021a}. We hope that our hybrid approach will ultimately create foundation models useful for as-yet-unknown galaxy morphology tasks.

\section{Data}


\label{sec:data}

The key properties of telescope images are angular resolution (how sharp sources appear) and depth (how bright sources must be to distinguish themselves from the background). Images are grouped into surveys according to the telescope used and the operating conditions. Galaxies identified in those surveys may then be labelled by Galaxy Zoo volunteers - members of the public answering a series of questions about the features of each galaxy via a browser. Below, we briefly summarise the properties of the surveys from which draw our images and the Galaxy Zoo campaigns (project versions) that labelled them.


Galaxy Zoo 2 (GZ2, \citealt{Willett2013}) labelled images from the Sloan Digital Sky Survey DR7 \citep{Abazajian2009}. This survey was the first to which deep learning was applied \citep{Dieleman2015} and remains a common reference dataset. GZ2 volunteers were asked whether galaxies had features including bars, bulges, and spiral arms. We include the 209,239 galaxies in the GZ2 main sample. 

Galaxy Zoo Legacy Survey (LegS, Walmsley et al. in prep.) labels images from the DESI Legacy Imaging Surveys \citep{Dey2018}. The individual surveys are DECaLS, MzLS, and BASS, each producing comparable images. LegS images are deeper than those in GZ2, revealing fainter galaxy details.
Most (305k) of our volunteer labels for this campaign are drawn from Galaxy Zoo DECaLS \citep{Walmsley2022decals} which labelled a non-random subset of DECaLS images. The remainder (60k) are DECaLS images labelled subsequently. Volunteers were asked similar questions to GZ2, with an additional focus on fainter details (e.g. is the bar strong or weak, is the galaxy disturbed). We also include the 1.34 million nearby (redshift $z < 0.1$, \citealt{Zou2019}) galaxies not labelled by volunteers as a large unlabelled subset of well-resolved galaxies for contrastive learning. 

Galaxy Zoo Hubble \citep{Willett2017a} and Galaxy Zoo CANDELS \citep{Simmons2017} labelled galaxy images from Hubble Space Telescope surveys \cite{Grogin2011, Griffith2012}. Despite the improved resolution of a space-based telescope, the target galaxies are far more distant (redshifts $1 \leq z \leq 3$ vs. $z \leq 0.4$ for LegS and $z \leq 0.15$ for GZ2) and hence appear less well-resolved (obscuring fine details) and relatively faint. However, these distant galaxies appear earlier in their evolution and show unique features such as starforming clumps. Volunteers were asked new questions about these clumps and fewer questions about detailed morphology.  We include the 100k and 50k galaxies from the GZ Hubble and GZ CANDELS primary samples, respectively.

For our downstream task, we use volunteer labels from Galaxy Zoo Rings (Walmsley et al. in prep.). Volunteers labelled a selected subset of LegS galaxies as being ringed or not using the Galaxy Zoo Mobile swipe interface. 10 volunteers label each image and the majority response is used as a binary label. This is a new task - Galaxy Zoo volunteers do not label galaxies as ringed in any of our pretraining campaigns. Rings includes 93k galaxies and is roughly balanced between rings and non-rings. 

We group the 552k galaxies labelled during the campaigns above with the 1.34m unlabelled nearby LegS galaxies and name the aggregate dataset \textbf{GZ-Evo v0.1}. GZ-Evo is the largest collection of volunteer galaxy labels to date, with 96.5 million volunteer answers. GZ-Evo is publicly available at \url{www.github.com/mwalmsley/pytorch-galaxy-datasets}.




\section{Method}
\label{sec:method}

GZ-Evo has two properties that motivate our approach. 
First, the questions volunteers were asked vary between campaigns according to the science goals of the organisers and the nature of the images (Sec. \ref{sec:data}). Even where questions do not change, subtle details like the tutorial and the answer icons cause significant distribution shifts \citep{Walmsley2022decals}. We therefore choose to treat answers as non-exchangeable between campaigns. Across GZ-Evo, there are 206 possible answers to 65 questions. We hypothesize that learning to jointly predict all answers may help build a representation that generalizes to new tasks, and do so using a custom Dirichlet loss function (Sec. \ref{sec:dirichlet_loss}).

Second, as often in astronomy, we have plentiful (1.34m) unlabelled images. We would like to exploit these images to help solve downstream tasks while also using the substantial (552k) set of labelled images. We therefore introduce a hybrid pretraining/contrastive approach (Sec. \ref{sec:adapting_byol}).

\subsection{Multi-Question Multi-Campaign Dirichlet Loss}
\label{sec:dirichlet_loss}

Our aim to learn from volunteers answering different questions on different images, where any particular image has only a small subset of possible questions answered.

\citet{Walmsley2022decals} introduced the Dirichlet loss function in the context of managing heteroskedastic votes for questions in the same Galaxy Zoo campaign.

\begin{equation}
    \label{multivariate_per_q_likelihood}
    \mathcal{L}_q = \int \text{Multi}(k|\rho, N) \text{Dirichlet}(\rho| \alpha) d\alpha
\end{equation}

where, for some target question $q$, $k$ is the (vector) counts of responses (successes) of each answer, $N$ is the total number of responses (trials) to all answers, and $\rho$ is the (vector) probabilities of a volunteer giving each answer. $\rho$ is drawn from $\text{Dirichlet}(\rho|\alpha)$, where the model predicts the Dirichlet concentrations $\alpha$. Multinomial and Dirichlet distributions are conjugates and hence the integral is analytic. Intuitively, this loss corresponds to the odds of observing $k$ heads (votes for an answer) after $N$ coin flips (volunteers asked) assuming a (model-predicted) distribution for the bias of that coin. The Dirichlet-Multinomial distribution allows models to flexibly express uncertainty through wider posteriors and confidence through narrower posteriors.

Assuming answers to different questions are independent, the loss may be applied to multiple questions via the sum
\begin{equation}
    \log \mathcal{L} = \sum_q \mathcal{L}_q(k_q, N_q, f^w_q)
\end{equation}
where, for question $q$, $N_q$ is the total answers, $k_q$ is the observed votes for each answer, and $f^w_q$ is the values of the output units corresponding to those answers (which we interpret as the Dirichlet $\alpha$ parameters in Eqn. \ref{multivariate_per_q_likelihood}).

 To extend to multiple campaigns, we note that this loss naturally handles questions with no answers as $p(a=0|\alpha, N=0)=1$ for all $\alpha$ and hence $\frac{\partial \mathcal{L}}{\partial \alpha} = 0$, meaning unanswered questions do not affect the training gradients. We can therefore construct a multi-campaign vote count vector $K$ where $K_i$ is the votes for answer $i$ and $i$ indexes all answers across all questions \textit{across all campaigns}. For a galaxy labelled in any single campaign, $K_i$ is 0 for any answer $a_i$ to any question not asked in that campaign. Every answer is always predicted but the prediction only affects training if votes for that answer are non-zero. Intuitively, this corresponds to having zero recorded votes to questions not asked. Questions are effectively treated as orthogonal prediction tasks using the same representation.

\subsection{Adapting BYOL with Supervised Projection Head}
\label{sec:adapting_byol}

We would like to guide the contrastive learning process using Galaxy Zoo labels where available (via our loss function above).
Unfortunately, all supervised contrastive frameworks we identified \citep{Khosla2020, Tian2020,Wang2021a,Zhao2022} assume a classification context, often mirroring the positive/negative paired views concept of e.g. \citet{Sohn2016} with a contrastive supervised loss that rewards/penalizes similar representations for pairs of the same/different class. 
Our volunteer responses are of the form $k$ of $N$ volunteers gave answer $a$ to question $q$ and are not easily reduced to e.g. majority class labels due to heteroskedasticity (typically $5 < N < 40$) and the overlapping nature of the questions.
We instead use BYOL \citep{Grill2020}, an unsupervised contrastive framework which relies only on views of the same object, and extend it to a supervised context.

Briefly, BYOL consists of two encoders, where the target encoder $f_\xi$ is an exponential weighted average of the online encoder $f_\theta$. Each encoder projects strongly-augmented image views to representations $y_\xi$ and $y_\theta$ and then (via fully-connected heads) to projections $z_\xi$ and $z_\theta$. The online encoder has a further fully-connected head $q_{\theta\text{, con}}$ which predicts the projection of the target encoder $q_{\theta\text{, con}}(z_\theta) = \hat{z_\xi}$.
To train, the parameters $\theta$ of the online encoder are updated to best predict the representation of the target encoder when both encoders are fed different strongly-augmented views of the same image. The parameters $\xi$ of the target encoder are then updated to the new rolling average of $\theta$.

To guide the representation learned by BYOL, we add a new fully-connected prediction head to the online encoder. This head $q_{\theta\text{, sup}}(y_\theta) = \hat{K}$ predicts the volunteer votes $K$ from the online-encoded representation, analogously to the final layers of a CNN in the standard supervised context. The head outputs are constrained via sigmoid activation to the range (1, 100) such that they may be interpreted as the parameters of a Dirichlet distribution (Sec. \ref{sec:dirichlet_loss}).

 We combine the contrastive prediction head loss with the new supervised prediction head loss via a normalised weighted sum, with hyperparameter $\lambda$ weighting contrastive invariance and supervised predictive performance. 
 
\begin{equation}
    \label{eqn:combined_loss}
    \mathcal{L} = \mathcal{L}_\text{con}(q_{\theta\text{, con}}(z_\theta, z_\xi) + \lambda \mathcal{L}_\text{Dirichlet}(q_{\theta\text{, sup}}(y_\theta), K)/\mathcal{L}_\text{con}
\end{equation}

\subsection{Training and Evaluation Details}

Our aim is to learn a representation that maximises performance on the downstream task (classifying galaxies as ringed or not-ringed, with labels from Galaxy Zoo Mobile, Sec. \ref{sec:data}), without using downstream labels until the representation is trained.

We experiment with three representation-learning approaches on GZ-Evo: purely-contrastive BYOL (`Contrastive'); purely-supervised pretraining on all GZ-Evo answers with our Dirichlet loss function (`Pretrained'); and our hybrid contrastive/supervised approach (`Hybrid', BYOL with the added supervised prediction head and Dirichlet loss function). For each experiment, we evaluate the quality of our learned representation by training a linear classifier to solve the downstream rings task on that (now frozen) representation.
Our baseline is directly training a supervised model on the downstream labels (`Direct').

The Pretrained and Direct classifiers use Zoobot \cite{Walmsley2022}, a version of EfficientNetB0 \citep{Tan2019a} with the Dirichlet loss above and standard (as opposed to contrastive-style) augmentations of flips, rotations, and non-central crops. Pretrained learns from all 206 GZ-Evo answers and the representation is used as input features for the linear classifier, while Direct is directly trained on only the ring labels.
For BYOL's encoder (with or without the Dirichlet supervised head), we experiment with EfficientNetB0 and ResNet18 \citep{He2016}.

All models use Adam \citep{Kingma2015} with standard hyperparameters (learning rate 0.001, $\beta_1=0.9$, $\beta_2=0.999$). We divide GZ-Evo into fixed 70/10/20 train/val/test sets (for rings, 50/50 train/test) and reserve the test sets for future work. We evaluate downstream linear accuracy with 5-fold validation on the ring train set. All models are trained to convergence, as measured by supervised validation loss (on rings, for direct training, or on GZ-Evo, for pretraining) or contrastive loss. All models use color images downscaled to 128x128 pixels.

\section{Results \& Discussion}
\label{sec:Results}

\begin{figure}
    \centering
    \includegraphics[width=\columnwidth]{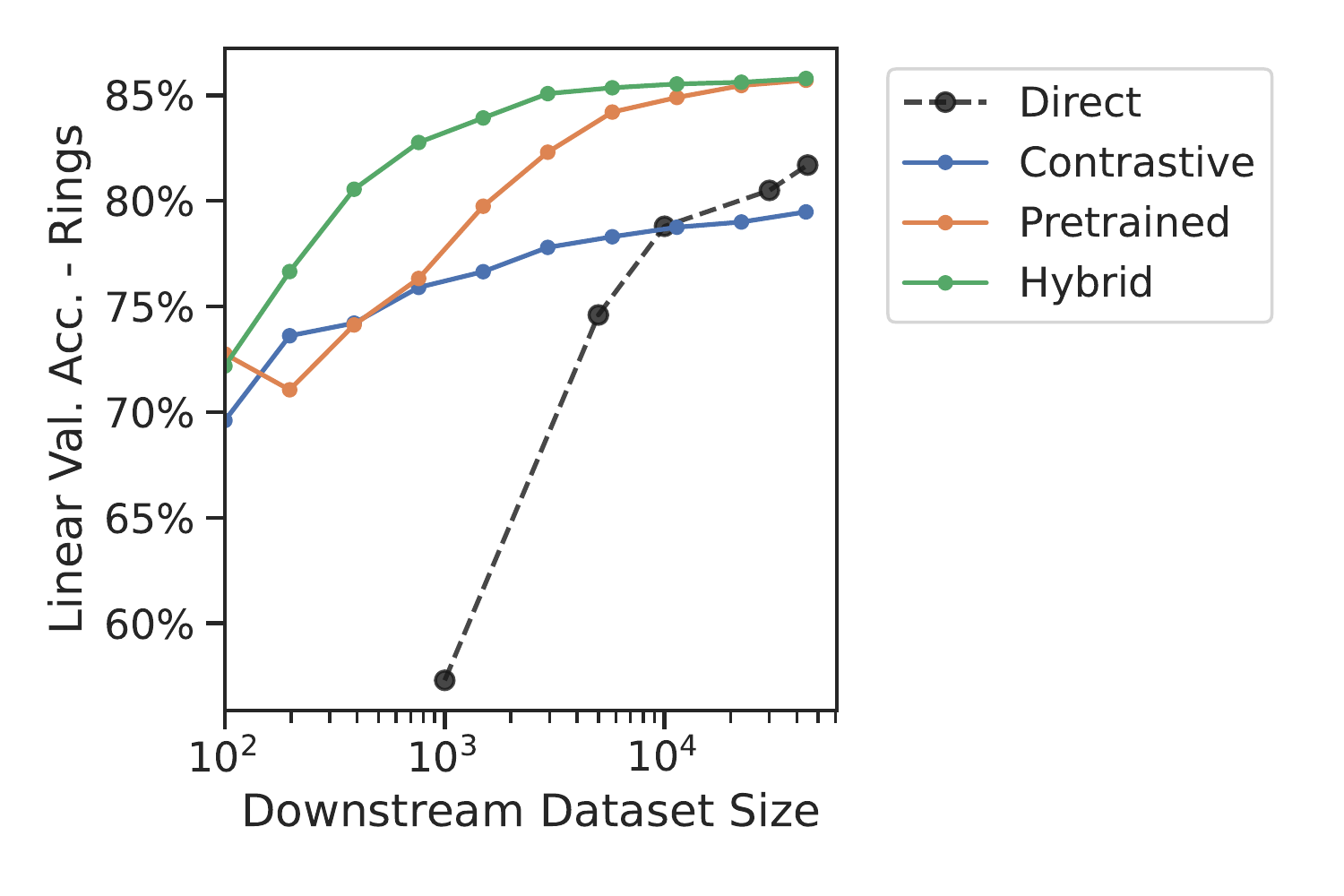}
    \caption{For our downstream task, classifying ring galaxies, our hybrid pretrained/contrastive approach matches or outperforms purely-pretrained and purely-contrastive approaches at all downstream dataset sizes.}
    \label{fig:downstream_val_acc}
\end{figure}

Our hybrid pretrained/contrastive approach matches or outperforms purely-contrastive and purely-pretrained approaches on all downstream ring dataset sizes (Fig \ref{fig:downstream_val_acc}). In the low-label transfer regime (761 downstream labels), our hybrid approach achieves a 6\% accuracy increase (relative error reduction of 28\%) over pretraining or contrastive learning alone. As additional downstream labels are collected, hybrid training provides consistent but decreasing improvements, with pretraining eventually matching hybrid learning at the limit of our dataset size (44k). Both methods outperform direct training. Purely-contrastive learning provides a strong baseline, outperforming direct training up to approx. 7k downstream labels, but is not competitive with pretraining. 




\begin{figure}[t]
    \centering
    \includegraphics[width=\columnwidth]{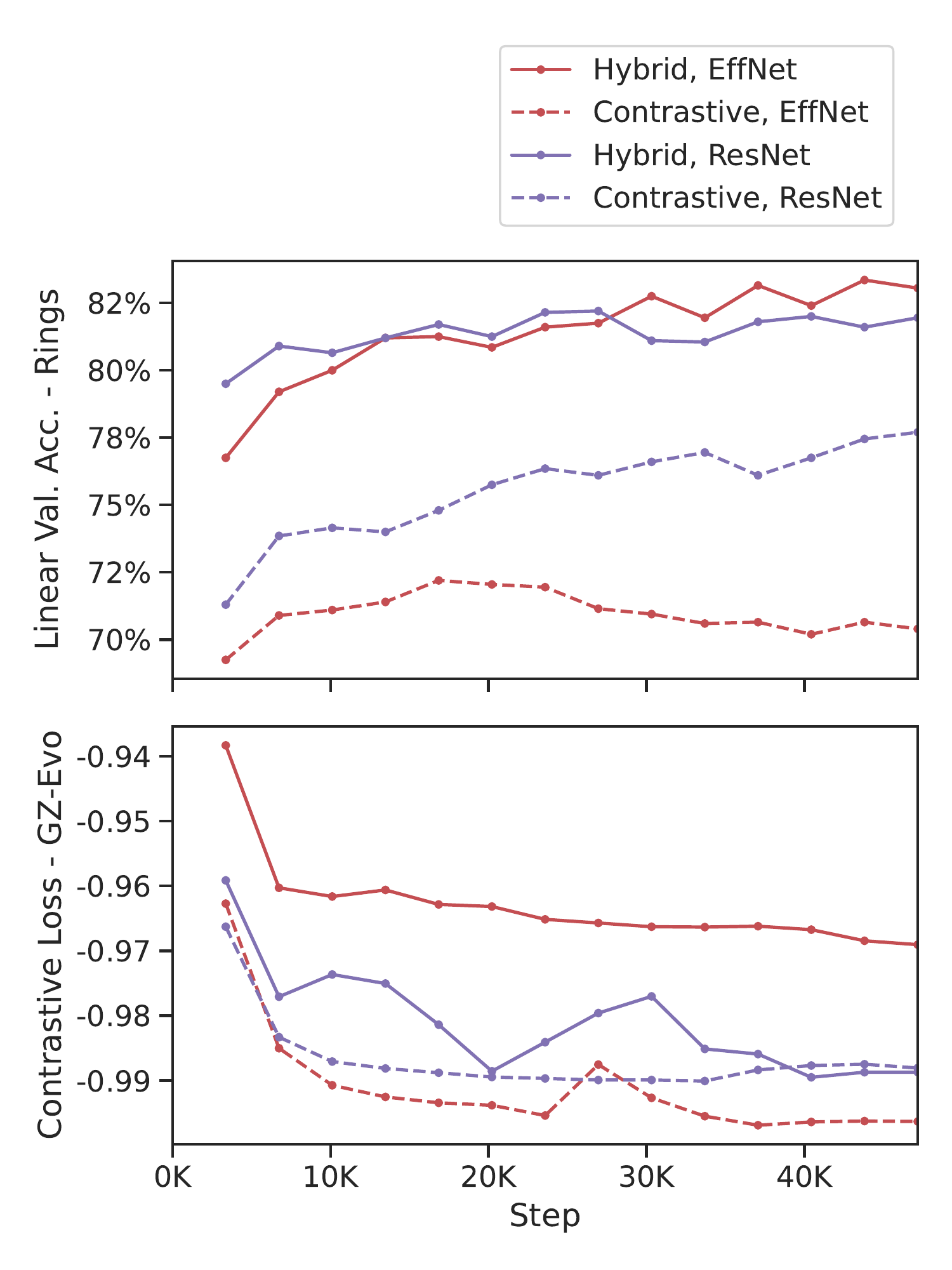}
    \caption{The representation learned by our hybrid pretrained/contrastive approach achieves higher downstream accuracy when classifying ring galaxies than a purely-constrastive approach (BYOL). The purely-contrastive approach achieves a lower (better) contrastive loss but \textit{worse} downstream performance - particularly when using EffNetB0 as an encoder. With the hybrid approach, ResNet18 performs similarly to EffNet.}
    \label{fig:performance_overview}
\end{figure}

Fig. \ref{fig:performance_overview} compares the downstream accuracy and contrastive loss of the purely-contrastive vs. hybrid approach. The purely-contrastive approach better solves the contrastive problem (achieving a lower contrastive loss) but the learned representation is worse for solving the downstream task. Considering the encoder, downstream accuracy is particularly disconnected from contrastive loss with EfficientNetB0 - unless our hybrid second head is added, in which case downstream accuracy is similar to ResNet18. This emphasizes that good contrastive solutions do not necessarily imply good downstream accuracy, and that hybrid supervised training can help guide those solutions.


\section{Conclusion}

New galaxy morphology tasks are often related to earlier tasks for which labels have already been collected. We experiment with pretraining on these labels, in combination with contrastive learning, to solve new downstream tasks. We first introduce GZ-Evo, a collection of 96.5M volunteer responses on the morphology of 552K galaxy images with a further 1.34M comparable unlabelled images. The responses span a wide range of related prediction tasks - there are 206 possible telescope/instruction/answer combinations, but any given galaxy only has responses for a minority of these. We jointly learn to predict all tasks using a Dirichlet loss function that naturally handles missing responses. Pretraining EfficientNetB0 on all GZ-Evo tasks then applying a simple linear classifier to the resulting representation outperforms direct training on our downstream task (identifying ringed galaxies), even with 44k downstream labels. 

We then add a new supervised prediction head to the BYOL contrastive learning framework, allowing us to learn from both the supervised pretraining task (where GZ-Evo labels exist) and the standard contrastive invariance loss (all galaxies). Our hybrid approach matches or outperforms purely-pretraining or purely-contrastive approaches, particularly in the low-labels regime (approx. 750 downstream labels). 

We hope that this simple hybrid method will help astronomers leverage the last decade of volunteer effort to answer new science questions.

\section{Acknowledgements}

This work was made possible by the Galaxy Zoo volunteers, who collectively created the crucial morphology labels of GZ-Evo (and much, much more). Their efforts are individually and gratefully acknowledged \href{http://authors.galaxyzoo.org}{here}. Thank you.

We thank the anonymous reviewers whose comments improved this work.

MW, IVS, MB and AMS gratefully acknowledge support from the UK Alan Turing Institute under grant reference EP/V030302/1. IVS gratefully acknowledges support from the Frankopan Foundation. 

\bibliography{references}
\bibliographystyle{icml2022}

\end{document}